# Artificial Neural Network Fuzzy Inference System (ANFIS) For Brain Tumor Detection

Minakshi Sharma[1], Dr. Sourabh Mukharjee[2]

**ABSTRACT**

Detection and segmentation of Brain tumor is very important because it provides anatomical information of normal and abnormal tissues which helps in treatment planning and patient follow-up. There are number of techniques for image segmentation. Proposed research work uses ANFIS (Artificial Neural Network Fuzzy Inference System) for image classification and then compares the results with FCM (Fuzzy C means) and K-NN (K-nearest neighbor). ANFIS includes benefits of both ANN and the fuzzy logic systems. A comprehensive feature set and fuzzy rules are selected to classify an abnormal image to the corresponding tumor type. Experimental results illustrate promising results in terms of classification accuracy. A comparative analysis is performed with the FCM and K-NN to show the superior nature of ANFIS systems.

## 1. INTRODUCTION

The classification of MR images is becoming increasingly important in the medical field since it is crucial for treatment planning and diagnosing abnormality (For e.g Brain Tumor),measure tissue volume to see tumor growth, study anatomical structure and patient follow up. Manual classification of magnetic resonance (MR) brain tumor images is a challenging and time-consuming task [1].Manual classification is highly prone to error due to interobserver variability and human error. As a result, the classification results are highly inferior which leads to fatal results. Thus, an automatic or semi-automatic classification method is highly desirable as it reduces the load on the human observer , large number of cases can be handled with same accuracy, also,results are not affected due to fatigue, data overload, faster communication .There are no universal algorithm for segmentation of every medical images. Different body parts MRI image needs different type of segmentation. Various methods proposed in the literature have met with only limited success [3,4] due to overlapping intensity distributions of healthy tissue, tumor, and surrounding edema. The most common class of methods is statistical classification using multiparameter images [5]. These methods are highly intensity based and hence the accuracy is very low. Warfield et al. [6] combined elastic atlas registration with statistical classification. Marcel Prastawa [7] used a modified spatial atlas for classification which includes prior probabilities for tumor and edema. Another group of researchers highly depend on computational intelligence for MR brain tumor classification which guarantees high accuracy. Zumray et.al [8] elaborates the inferior results of multilayer perceptron for the biomedical image classification problem. The Self Organizing Feature Map (SOFM) ANN based algorithms [9] shows excellent results in the classification of brain tumor images. Other studies based on learning vector quantization (LVQ) ANN show the potential of these architectures in the case of supervised classification. Hopfield neural networks (HNN) [10] prove to be efficient for unsupervised pattern classification of medical images, particularly in the detection of abnormal tissues. The use of ART2 network for pattern recognition has been studied by Solis and Perez [11]. Several modifications on the existing neural networks are implemented successfully and superior results have been achieved. One such work is reported by William Melson [12]. Besides being robust, they require large training dataset to achieve high accuracy. This increases the dimensionality problem which accounts for the complexity of the model. On the other hand, several researches based on fuzzy logic techniques are also reported in the literature. Marcin Denkowski [13] used rule based fuzzy logic inference for MR brain image classification. Experiments based on fuzzy C-means algorithm are also proposed in the literature [14].Yang and Zheng [15] implemented a modified fuzzy C-means algorithm for image classification. The fuzzy set theoretic models try to mimic human reasoning and the capability of handling uncertainty, whereas the neural network models attempt to emulate the architecture and information representation schemes of the human brain. Integration of the merits of the fuzzy set theory and neural network theory promises to provide, to a great extent, more intelligent systems to handle real life problems. A neuro-fuzzy approach as a combination of neural networks and fuzzy logic has been introduced to overcome the individual weaknesses and to offer more appealing features. The ultimate goal of applying such a system is to get rid of imprecise information present in an image such as

[1].Minakshi sharma is working as Assistant Professor in the Department of IT in GIMT kanipla, kurukshetra, India. pursuing PhD

[2].Dr.sourabh mukharjeeis working as Professor in the Department of Computer science in Banasthali university,rajasthan,

pixel greyness ambiguity, geometrical segmentation of the image and the uncertain interpretation of a scene. This exploits, respectively, the learning capabilities and the descriptive power of systems, thus providing results characterized by a high interpretability and good degree of accuracy [16]. Segmentation of images using neuro fuzzy model has been studied by Rami J. Oweis and Muna J. Sunna [17] .Image segmentation using neuro fuzzy tools are also implemented by Mausumi Acharyya [18]. ANFIS is one of the widely used neuro-fuzzy systems.In this work, the neuro-fuzzy based approach namely adaptive neuro fuzzy inference system (ANFIS) is used for MR brain tumor classification.

## 2. Proposed Methodology

The methodology used for MR brain tumor images is Divided in to four steps and third step is further divided in to four parts as shown in fig. 2.1 and 2.2.

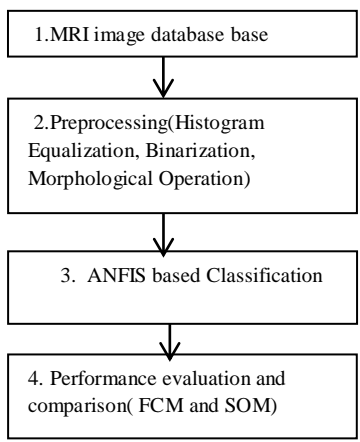

Figure 2.1: Proposed Methodology for Classification of Brain Tumor.

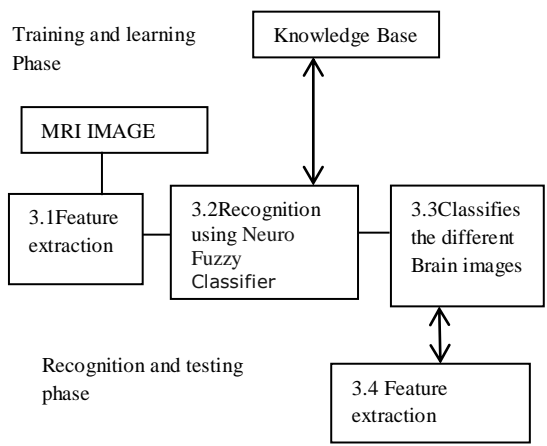

Figure 2.2: Proposed Methodology for ANFIS based brain tumor classification

**2.1 MR Image database**: MR image database consists astrocytoma, giloma type of brain tumor images of GRADE I to IV. These images are collected from web resource - http://mouldy.bic.mni.mcgill.ca/brainweb/

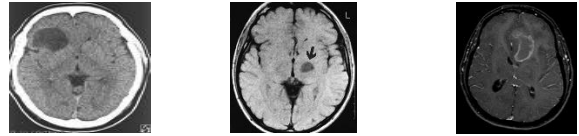

Figure 2.3 Sample Data Set

### 2.2. Image Preprocessing

Image preprocessing technique represents essential step of image segmentation which has a great impact on subsequent steps. In the proposed work ,three preprocessing techniques are used. They are-

**a)Histogram Equalization**

Histogram equalization is the technique by which the dynamic range of the histogram of an image is increased. Histogram equalization assigns the intensity values of pixels in the input image such that the output image contains a uniform distribution of intensities. It improves contrast of an image.

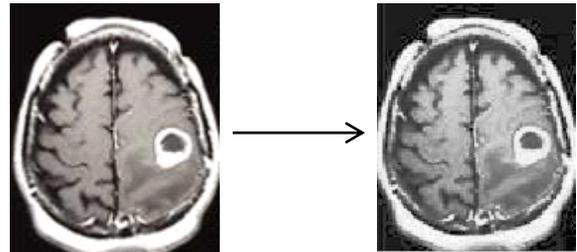

Fig 2.4 Histogram Equalized Image

**b)Binarization**

Image binarization converts grey scale image in to a binary image(either black or white) based on some threshold value.

$$G(x,y)=\begin{cases}1 & f(x,y) \geq T \\ 0 & f(x,y) < T\end{cases} \quad (1)$$

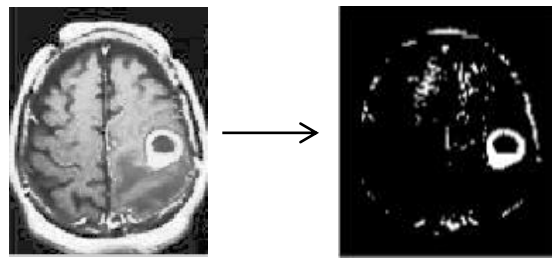

Figure 2.5 Binarized image for the given grey scale image

In the proposed work only one threshold value is chosen for the entire image which is based on

intensity histogram (mean of intensity values are taken)

c) Morphological Operations

This is used as a image processing tools for sharpening the regions and filling the gaps for binarized image. There are four basic morphological operations are defined like dilation, erosion, opening and closing. Here, proposed work uses only dilation and erosion. In erosion every pixel that touches background pixel is changed in to background pixel. Erosion makes object smaller and break single object in to multiple objects. Mathematically erosion can be represented as,

(A⊖B)(x)={x∈ X, x=a+b: a ∈A b∈B}  (2)

Where A represents matrix of binary image and B represents mask. Whereas, dilation change background pixel that touches object pixel is changed in to object pixel. Dilation Merges multiple objects in one. . Mathematically dilation can be represented as,

(A⊕B)(x)={x∈ X, x=a+b: a ∈A b∈B}  (3)

2.3 Feature Extraction

Features are the characteristics of the objects of interest in an image. Feature extraction is the technique of extracting specific features from the pre-processed images of different abnormal categories in such a way that the within – class similarity is maximized and between – class similarity is minimized. Earlier research works report many feature extraction techniques. In this proposed work Gray Level Co- occurrence Matrix (GLCM) features are used to distinguish between normal and abnormal brain tumors. GLCM is the gray-level co-occurrence matrix (GLCM), also known as the gray-level spatial dependence matrix. By default, the spatial relationship is defined as the pixel of interest and the pixel to its immediate right (horizontally adjacent).GLCM has following features: Autocorrelation, Contrast, Correlation, Cluster Prominence, ClusterShade, Dissimilarity, Energy, Entropy, Homogeneity, Maximum probability , Sum of squares, Sum average, Sum variance, Sum entropy, Difference variance, Difference entropy, Information measure of correlation, information measure of correlation, Inverse difference normalized. Out of these 19 features 7 features are taken.

**1. Contrast:** It Returns a measure of the intensity contrast between a pixel and its neighbor over the whole image. Contrast is 0 for a constant image.

Contrast= $\sum_{i,j} |i-j|^2 \, p(i,j)$  (4)

Where, P(I,j) pixel at location (i,j)

**2. Angular Second Moment (ASM):** It provides a strong measure of homogeneity.

ASM=$\sum_{i,j} p^2 \, (i,j)$  (5)

**3. Homogeneity (HOM):** Returns a value that measures the closeness of the distribution of elements in the GLCM to the GLCM diagonal.

HOM=$\sum_{i,j} \frac{p(i,j)}{1+|i-j|}$  (6)

**4. Inverse Difference Moment (IDM):** It is the measure of local homogeneity.

IDM= $\sum_i \sum_j \frac{1}{1+(i-j)^2} \, p(i,j)$  (7)

**5. Energy (E):** Returns the sum of squared elements in the GLCM. Energy is 1 for a constant image.

E=$\sum_{i,j} p(i,j)^2$  (8)

**6. Entropy (EN):** It is a measure of randomness.

EN=$\sum_{b=0}^{L-1} p(i,j) \log_2\{p(i,j)\}$  (9)

Where, L is no. of different values which pixels can adopt.

**7. Variance (VAR):** It is the measure that tells us by how much the gray levels are varying from the mean.

VAR=$\sum_i \sum_j p(i,j)p(i,j) - \mu^2$  (10)

In the proposed work , seven GLCM features are calculated per image in four directions 0,45,95 135 and hence the number of input linguistic variables are seven. The number of output linguistic value is equal to the number of patterns used in this work.Table1 show a sample of features value for image 1 and image 2.Based upon this value normal and abnormal brain can be differentiated.

|    | Features | IMAGE1 Range(High-Low) | IMAGE2 Range(High-Low) |
|----|----------|------------------------|------------------------|
| 1. | Contrast | 7.08e+00-6.98e+00 | 3.60e+00-4.53e-001 |
| 2  | ASM | 8.76e-001-8.72e-001 | 6.05e-001-6.72e-001 |
| 3  | HOM | 8.87e-001-8.62e-001 | 8.72e-001-8.62e-001 |
| 4  | E | 2.93e-001-2.85e-001 | 2.28e-001-2.26e-001 |
| 5  | EN | 2.68e-001-3.44e-001 | 2.72e-001-3.01e-001 |
| 6  | VAR | 8.96e-001-8.54e-001 | 9.06e-001-8.81e-001 |
| 7  | IDM | 9.92e-001-9.90e-001 | 9.94e-001-9.93e-001 |

Table1 Seven features with range (low and High) of image1 and image2

A sample of fuzzy if-then rules framed for the MR brain tumor classification is shown below :

**Rule 1:** If a is contrast1 and b is correlation1 and c is energy1and d is entropy1 and e is IDM1 and f is variance1,then output = 1

**Rule2:**If a is contrast2 and b is correlation2 and c is energy2and d is entropy2 and e is IDM2 and f is

variance2, then *output = 2*

**Rule3:** If a is contrast3 and b is correlation3 and c is energy3 and d is entropy3 and e is IDM3 and f is variance3, then output = 3 The number of membership functions used in this work is 2 (low and high) and hence there are 49 rules framed for this image classification system. These fuzzy if-then rules form the input for the ANFIS architecture.

### 2.4 ANFIS Architecture

The structure of ANFIS consists of 7 inputs and single output. The 7 inputs represent the different textural features calculated from each image. Each of the training sets forms a fuzzy inference system with 16 fuzzy rules. Each input was given two generalized bell curve membership functions and the output was represented by two linear membership functions. The outputs of the 49 rules are condensed into one single output, representing that system output for that input image.

The data set is divided into two categories: training data and testing data. The training data set consists of images from all the four tumor types. These training samples are clustered into four different regions namely white matter, grey matter, cerebrospinal fluid and the abnormal tumor region using the fuzzy C-means (FCM) algorithm. The cluster center of the tumor region for all the four classes are observed and stored. In the testing process, features are extracted and match with the best possible solution.

### 2.5 Training Algorithm

There are several training algorithms for feed forward networks. The gradient is determined using a technique called back propagation, which involves performing computational backwards through the network. The simplest implementation of back propagation learning adjusts the network weights in the direction in which the performance function decreases more rapidly. The algorithm used in this work is extracted from [25].

### 2.6 Performance measures

All classification result could have an error rate and on occasion will either fail to identify an abnormality, or identify an abnormality which is not present. It is common to describe this error rate by the terms true and false positive and true and false negative as follows: [8]

*True Positive (TP)*: the classification result is positive in the presence of the clinical abnormality.
*True Negative (TN)*: the classification result is negative in the absence of the clinical abnormality.
*False Positive (FP)*: the classification result is positive in the absence of the clinical abnormality.
*False Negative (FN)*: the classification result is negative in the presence of the clinical abnormality.

Sensitivity = TP/ (TP+FN) *100%
Specificity = TN/ (TN+FP) *100%
Accuracy = (TP+TN)/ (TP+TN+FP+FN)*100 %

| Algorithms | Sensitivity | Specificity | Accuracy |
|---|---|---|---|
| DWT+SOM[7] | 95.13 | 92.2% | 94.72 |
| DWT+PCA+KNN] | 96.2 | 95.3 | 97.2% |
| Second order+ANN | 91.42 | 90.1 | 92.22 |
| Texture Combined+ANN | 95.4 | 96.1 | 97.22 |
| Texture Combined+SVM | 97.8 | 96.6 | 97.9 |
| FCM | 96% | 93.3% | 86.6 |
| K-Mean | 80% | 93.12% | 83.3 |
| Proposed (ANFIS) | 96.6% | 95.3% | 98.67% |

Table 2: Comparison of classification performance for the proposed technique and recently other work

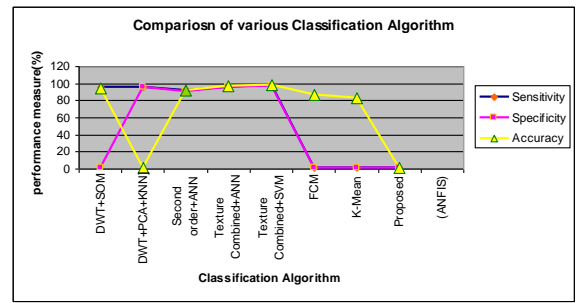

Fig 2.6 Comparison of classification performance for the proposed technique and recently other work

## 3. RESULTS

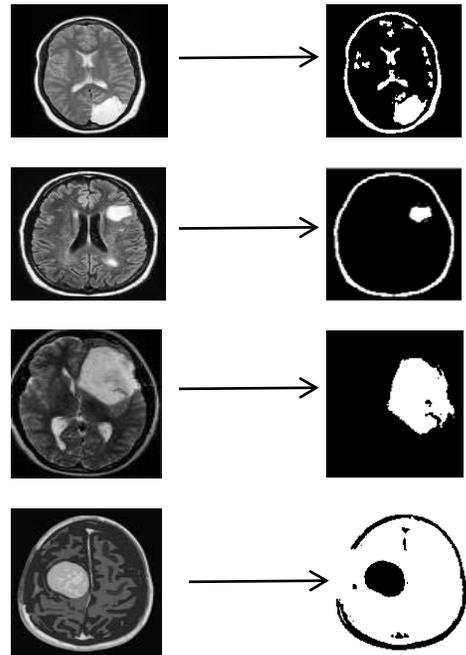

**Fig 3.1 Tumor Segmented from abnormal brain MRI image**

## 4. Conclusion

In this work, the application of ANFIS for MR brain tumor image classification is explored. Experimental results yield promising results for ANFIS as an image classifier. The classification accuracy of ANFIS as shown in fig. The future scope of this work is to enhance the ANFIS architecture to achieve high classification accuracy, measure thickness and volume of tumor